\documentclass[a4paper]{jpconf}
\usepackage{xcolor}
\usepackage{graphicx}
\usepackage[numbers,sort&compress,square]{natbib}
\newcommand{\jq}[1]{\textcolor{black}{#1}}

\begin{document}

\title{Adversarial Sensor Errors for Safe and Robust Wind Turbine Fleet Control}

\author{Julian Quick$^{1}$, Marcus Binder Nilsen$^{1}$, Andreas Bechmann$^{1}$, Tran Nguyen Le$^{2}$, and Pierre-Elouan Mikael Réthoré$^{1}$}

\address{$^1$Department of Wind and Energy Systems, Technical University of Denmark, Roskilde, Denmark}

\address{$^2$Department of Engineering Technology and Didactics, Technical University of Denmark, Ballerup, Denmark}

\ead{juqu@dtu.dk}

\begin{abstract}
Plant-level control is an emerging wind energy technology that presents opportunities and challenges. By controlling turbines in a coordinated manner via a central controller, it is possible to achieve greater wind power plant efficiency. However, there is a risk that measurement errors will confound the process, or even that hackers will alter the telemetry signals received by the central controller. This paper presents a framework for developing a safe plant controller by training it with an adversarial agent designed to confound it. This necessitates training the adversary to confound the controller, creating a sort of circular logic or ``Arms Race''. This paper examines three broad training approaches for co-training the protagonist and adversary, finding that an Arms Race approach yields the best results. 
Results indicate that the Arms Race adversarial training reduced worst-case performance degradation from 39\% power loss to 7.9\% power gain relative to a baseline operational strategy.
\end{abstract}

\vspace{1em}

\section{Introduction}
Wind turbines are autonomous agents that typically employ a maximum power point control strategy to maximize individual agents' power production. Wind farm wake effects create significant agent interactions, where the control strategy of one turbine can significantly impact the power production of downstream turbines. As wind energy technology matures, it is becoming increasingly recognized that fleet-level control strategies offer a more robust and profitable control model \cite{meyers2022wind}. 

Early wind farm fleet control strategies focused on open-loop control strategies that initiate agent actions based on inflow measurements. In recent years, there has been increased attention on closed-loop control strategies, where agent actions are dynamically set based on wind farm feedback. Recently, research has examined data-driven reinforcement learning (RL) for wind farm fleet control \cite{abkar2023reinforcement,goccmen2025data}. \jq{Huang and Zhao applied an offline reinforcement learning framework using a discriminator-generator approach to train a controller using a pre-computed set of large eddy simulation data generated using data-driven and simple rules-based controllers \cite{huang2025wind}. 
Mole et al. trained a soft-actor-critic RL controller using an LES simulator \cite{mole2025reinforcement}. 
Soler et al. apply a double-deep Q-learning RL approach to the control of a single wind turbine, simulated using blade-element-momentum theory couple with turbulent inflow \cite{soler2024reinforcement}. 
For a comprehensive review, we refer to Abkar et al. \cite{abkar2023reinforcement}.}

When designing wind farm control strategies, it may be useful to account for sensor errors to avoid overly optimistic projections \cite{quick2020wake}. While most sensors have instantaneous uncertainty estimates, it is rarely clear how these uncertainties are correlated in time. Relying on procedural noise generation approaches \jq{(e.g., independent Gaussian draws with a fixed episodic bias)}  may overlook more dangerous realizations of sensor error, masking actual risks. Additionally, there is a cybersecurity risk that malicious actors could hack into a wind farm system and alter sensed values to degrade system performance. 

This work heavily utilizes terminology from game theory to formalize these concerns. From this point forward, the \textbf{wind farm controller} is referred to as a \textbf{"protagonist"}: an agent that takes in state observations and outputs actions in a mathematically defined environment to maximize a reward function. We refer to \textbf{RL-injected measurement errors} as adversarial measurement errors created by an \textbf{"adversary"}, in the sense that the goal of the adversary is to confound the wind farm controller. 
\jq{There is an important concept of "catastrophic forgetting," where fine-tuning/hardening an agent against data-driven adversarial conditions yields reduced performance in nominal conditions \cite{mccloskey1989catastrophic}. Several approaches have been developed where zoos of adversarial RL agents are developed to confound and then harden future  RL controllers \cite{vinyals2019grandmaster} 
}

Taken together, these considerations naturally lead to the research question we aim to answer: \\



\noindent \textit{Does training, through interaction with an adversary agent, yield greater protagonist robustness than training with procedurally generated noise?} \\

This work addresses this question by first developing an open-source RL framework to create measurement-confounding agents. These adversary agents will be trained to inject measurement noise into the wind turbine fleet's sensors to minimize the wind farm's performance. The protagonist agents are not treated as static baselines: they are trained both independently and jointly in adversarial settings, enabling a direct assessment of how interaction-driven learning influences robustness.

Specifically, this paper presents the adversarial framework using the open-source WindGym package \cite{WindGym}. 
\jq{We explore three different approaches for co-evolving wind farm controller and adversarial sensor error agents. To the authors' best knowledge, this is the first work to apply adversarial agent training to wind farm sensor robustness.} 

\section{Methods}

\subsection{Simulation Environment}
This study employs the Dynamic Wake Meandering model implemented in the Dynamiks software \jq{\cite{dynamiks}} as the physics simulator. In the presented work, no additional turbulence is superimposed on the flow, allowing us to focus on the large-scale wake dynamics.  



\subsection{Protagonist Agent Definition}
The protagonist agent is trained to predict actions, defined as changes in turbine yaw offsets.

\subsection{Reward Function Definition}
The protagonist agent is trained to maximize the total power of the wind farm relative to a baseline greedy control strategy. At each time step $t$, the reward function is defined as
\begin{equation}
    r_t = \frac{\bar{P}_\mathrm{agent}\left(s_t, a_t\right)}{\bar{P}_\mathrm{baseline}\left(s_t\right)} - 1 \,,
\end{equation}
where:
\begin{itemize}
    \item $r_t$ is the reward at time $t$, 
    \item $P_\mathrm{agent}\left(s_t, a_t\right)$ is the power produced by a given agent at time $t$  averaged over the past 10 agent steps,
    \item $\bar{P}_\mathrm{baseline}\left(s_t\right)$ is the power produced the baseline no-steering case averaged over the past 10 agent steps,
    \item $s_t$ is the environment state at time $t$ (e.g., a vector of turbine-specific wind speed, direction, power, and yaw values), and
    \item $a_t$ are the actions taken by the agent at time $t$ (e.g., a vector of changes in yaw offsets)
\end{itemize} 
Note that, while the agent is working with corrupted observations, $s_t + \epsilon_t$ where $\epsilon_t$ is the measurement error at time $t$, the power produced by the agent is based on uncorrupted physics. \jq{Furthermore, the wind direction and yaw error are both added to the observed yaw positions in $s_t$}.

\subsection{Expert  Agent Definition}

The WindGym PyWake expert agent determines yaw offsets based on sensed wind direction and speed. \jq{The PyWake agent runs an internal simulation optimization tool to determine yaw offset setpoints}. Due to its construction, adversarial sensor errors can at most degrade \jq{the PyWake agent} performance to baseline (zero yaw offset) behavior, but cannot force counter-productive actions that would further reduce power production. 
This is accomplished by only confounding the PyWake\jq{/simulation} part of the controller: the turbines will be offset relative to the mean wind direction specified within the simulation inflow definition, even when the offsets computed by PyWake assume a completely incorrect wind direction. Note that the baseline controller of 0 yaw offsets makes a similar assumption about knowing the true direction. This represents a conservative yet realistic constraint, as real turbine controllers typically include safety limits preventing extreme misalignment. Alternative configurations could permit more severe degradation, \jq{where confounding the reference wind direction and corresponding yaw orientation can cause the turbines to turn all the way around,} but this would not reflect typical operational constraints. \jq{For example, if the wind comes from the east and there is a 90 degree wind direction error, the PyWake simulation will assume the wind direction is from the North, finding optimal yaw offsets based on this simulation. If these optimal yaw offsets are zeros, the controlled turbines will be set to zero yaw offset from the true wind direction.}

\subsection{Procedural Noise Definition}

To establish a baseline for robust control, we define a "procedural noise" environment representing standard sensor uncertainties without malicious intent. This model creates a noisy observation $\hat{x}(t)$ by combining instantaneous Gaussian noise with a persistent episodic bias:

\begin{equation}
    \hat{x}(t) = x_\mathrm{true}(t) + \eta(t) + \beta\,,
\end{equation}
where $\eta(t)$ are independent draws of a zero-mean normal distribution, $\mathcal{N}(0, \sigma_x)$, and the bias term, $\beta$, is constant in time and drawn from a uniform distribution $\mathcal{U}[-\beta_\mathrm{max},\beta_\mathrm{max}]$. Here, $\sigma_x$ is the standard deviation of the instantaneous measurement error in measurement signal $x$, and $\beta_\mathrm{max}$ is the maximum magnitude of the bias term.

\subsection{Adversary Agent Definition}

Instead of viewing adversarial attacks as purely malicious hacks, we can constrain the adversarial generation process to produce only physically plausible, real-world scenarios.

The advection of wakes makes this problem inherently time-dependent. The adversary introduces a measurement error, $\epsilon^k(t)$, to signal $k$. This error is defined so that $\epsilon^k(t=0)=0$, and the agent actions are defined as changes in each error term, $\Delta \epsilon^k(t)$. These changes are limited to one-tenth of the maximum allowable bias in each signal. The applied error is clipped if it exceeds these specified bounds. 

The adversary is trained under a strictly antagonistic objective, with its reward defined as the negative of the protagonist’s reward, thereby inducing a zero-sum interaction. The adversary and protagonist share the same observation space, but with an information asymmetry between them. The adversary observes the true state to compute sensor perturbations, while the protagonist observes only the confounded values resulting from those perturbations.

\subsection{Confronting the AI Arms Race Problem}

\begin{figure}[htb!]
    \centering
    \includegraphics[width=0.7\linewidth]{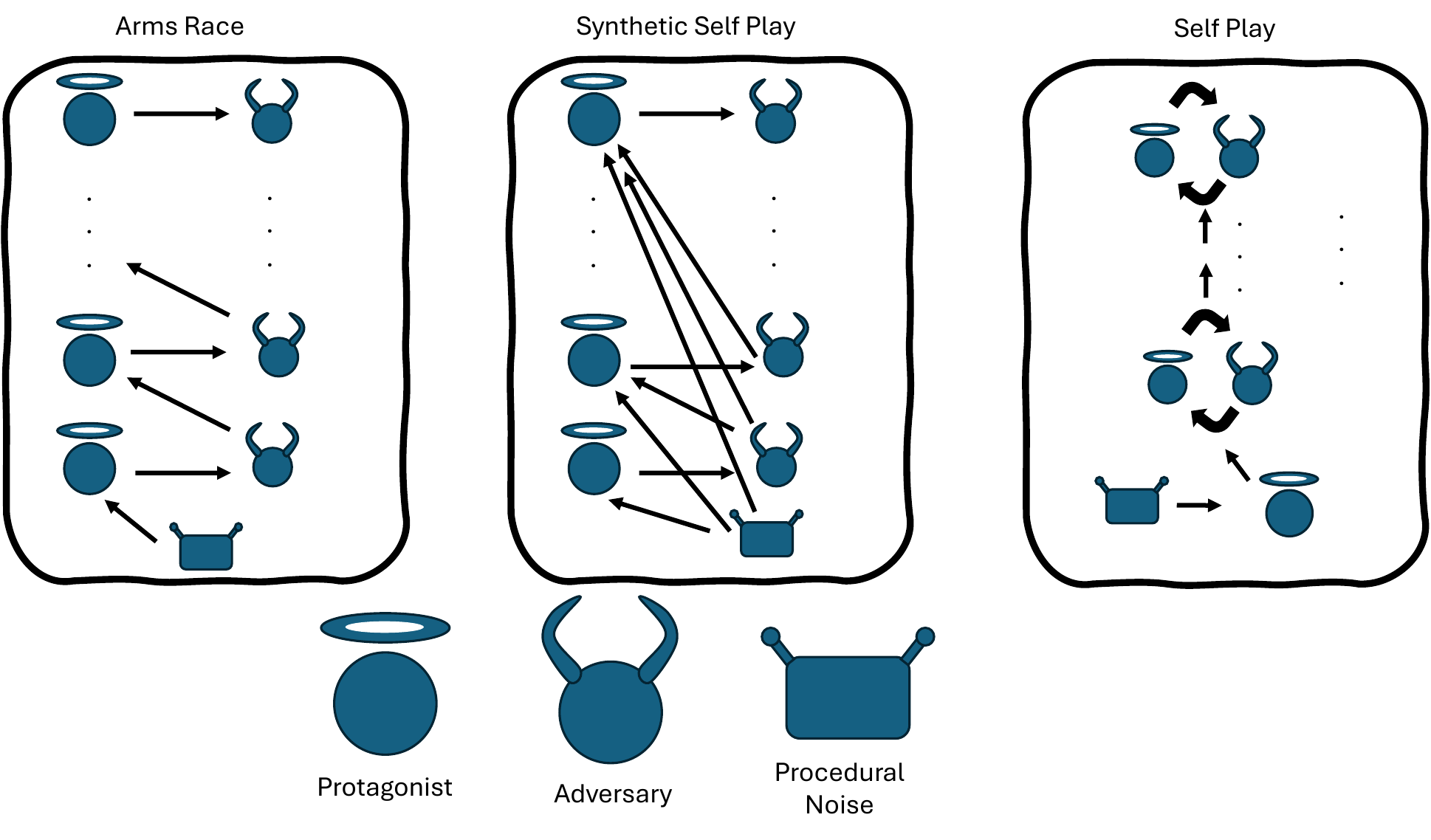}
    \caption{Training setups: Arms Race (left), Synthetic Self Play (middle), and Self-Play (right).}
    \label{fig:arms-race}
\end{figure}

We examine three training paradigms, illustrated in Figure \ref{fig:arms-race}. 
\begin{itemize}
    \item Left: The Arms Race is a sequential iteration strategy where protagonist $n$ trains exclusively against adversary $n-1$, which was trained against protagonist $n-1$, etc. Each generation discards all prior opponents. 
    \item Middle: Synthetic Self Play (SSP) has a similar setup to Arms Race, but protagonist N trains against a zoo of all prior adversaries ${0, 1, \dots, N-1}$, including procedural noise, sampling uniformly across the population at the start of each training episode. 
    \item Right: Self-Play is a concurrent optimization approach where the protagonist and adversary co-evolve within a single competitive loop, formulated as a zero-sum game using independent agent updates. 
    The Self-Play approach formulates the problem as a competitive zero-sum game and co-trains both agents simultaneously using the PettingZoo library \cite{terry2021pettingzoo}. Each Self-Play agent maintains separate actor and critic networks with independent optimizations.
\end{itemize}

\section{Application}

\begin{figure}[htb!]
    \centering
    \includegraphics[width=0.7\linewidth]{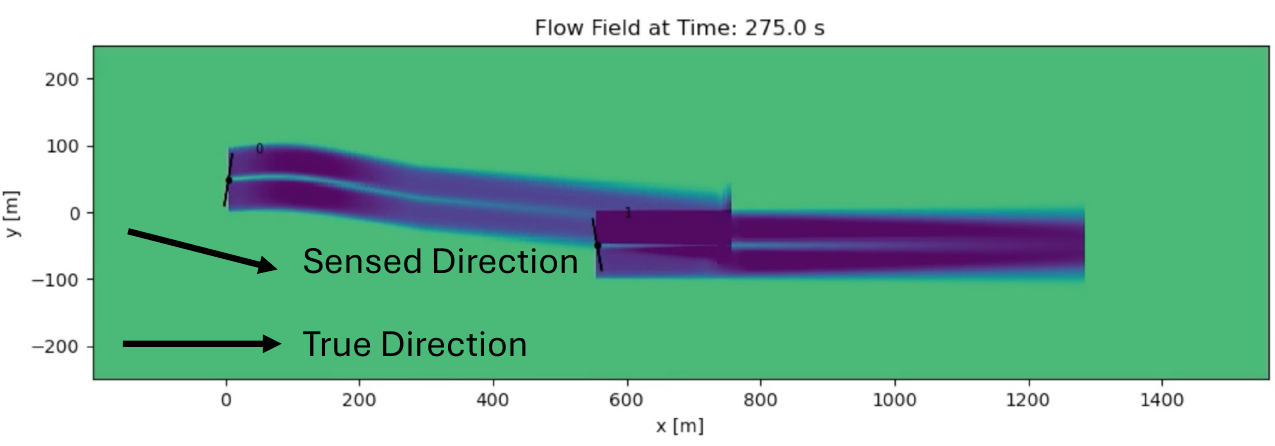}
    \caption{Snapshot of the wind farm flow field at hub height. Faster speeds are shown with brighter colors. 
    The yaw controller of the upstream wind turbine has been confounded to sense an incorrect wind direction.
    }
    \label{fig:setup}
\end{figure}

This study examines a wind farm consisting of two \jq{Vestas V80 \cite{pedersen2023PyWake}} turbines. This case captures essential wake interactions while providing a tractable environment for systematic analysis of different proposed training methodologies. The turbines are aligned in the East-West direction and spaced 7 rotor diameters apart. The inflow is limited to wind speeds between 6 and 7 m/s and wind direction between 267 and 273 degrees. \jq{These inflow conditions are sampled at the start of each episode using uniform distributions}. The Dynamic Wake Meandering model is run with time steps of 5 seconds, and the agents' actions are updated every 10 seconds. The flow is simulated without turbulence to simplify the problem. The basic computational experimental setup is visualized in Figure \ref{fig:setup}.

\jq{The protagonist and the adversary share the same state space: turbine-specific yaw, speed, direction, and power. The measured yaw of turbine $i$, $\hat{\gamma}_i$, is defined relative to the error in the measured wind direction, so errors in wind direction are also added to the yaw sensors: $\hat{\gamma_i} =\gamma_i + \epsilon_{\gamma_i} - \epsilon_{\theta_i}$. Note that there is a difference in sign because the yaw offset follows a right-hand-rule system while direction is defined using a left-handed orientation. The state space captures the current and the previous 10 values of each sensor. }




Signals are normalized to [-1, 1] using physical bounds: wind speed 0–30 m/s, direction 0–360$^\circ$, yaw offset ±45$^\circ$, and power 0–2 MW. Table \ref{tab:noise_parameters} summarizes the noise model parameters used for both the adversarial agent and the procedural noise environments.

\begin{table}[h]
\centering
\caption{Noise Model Parameters. The Max Bias ($\epsilon_{max}$) applies to both the Adversarial Agent and the Procedural Noise baseline. The Gaussian standard deviation ($\sigma$) applies only to the Procedural Noise baseline.}
\label{tab:noise_parameters}
\begin{tabular}{l | r r}
{Variable} & {Max Bias ($\pm$)} & {White Noise ($\sigma$)} \\
\hline
Wind Speed (m/s) & 4.0 & 0.5 \\
Wind Direction ($^{\circ}$) & 10.0 & 2.0 \\
Yaw Offset ($^{\circ}$) & 20.0 & 0 \\
Power (MW) & 0.5 & 0 \\
\end{tabular}
\end{table}


Both the protagonist and adversary agents are trained using the Proximal Policy Optimization (PPO) algorithm \cite{schulman2017proximal}. The hyperparameters used for training are standardized across the different experimental setups (Arms Race, SSP, and Self-Play) to ensure a fair comparison. The specific training parameters are detailed in Table \ref{tab:hyperparameters}. In each iterative training approach, the agents were trained for 250,000 timesteps per iteration. The Arms Race and SSP methods used 6 parallel environments, each with a rollout buffer of 512 steps. The Self-Play approach employed a single environment instance with a larger rollout buffer of 2048 steps.

\begin{table}[htb!]
    \centering
    \caption{PPO Hyperparameters used for training protagonist and adversary agents}
    \begin{tabular}{c|c|c}
         Parameter & Value & Description \\
         \hline
         Algorithm & PPO & Proximal Policy Optimization \\
         Learning Rate & $3 \times 10^{-4}$ & Step size for the optimizer \\
         Batch Size & 64 & Samples per gradient update \\
         Gamma ($\gamma$) & 0.99 & Discount factor \\
         Steps per Iteration & 250,000 & Timesteps per training round \\
    \end{tabular}
    \label{tab:hyperparameters}
\end{table}


The actor and critic networks for all agents were implemented as Multi-Layer Perceptrons with hyperbolic tangent activation functions. The output layer of the actor network determined the mean of a Gaussian distribution, while the log standard deviation was learned as a free, state-independent parameter. Across all training configurations (Arms Race, SSP, and Self-Play), the protagonist and adversary architectures utilized a network with two hidden layers of 128 units. 


The interaction between the agents and the wind farm simulation was discretized into fixed time steps. Environmental physics were simulated with a time step of 5 seconds. 
The agents observed the state and updated their control actions every 10 seconds. 

In each iterative training method (Arms Race and Self-Play), training was organized into "iterations." Each iteration consisted of 250,000 environmental time steps. Within these iterations, the PPO algorithm collected data in rollout buffers before performing gradient updates. The episode termination conditions and update frequencies were defined as follows:

\begin{itemize}
    \item Training Updates: In the Arms Race and SSP modes, the agents updated their networks every 512 time steps (collected across 6 parallel environments). In the Self-Play mode, updates occurred every 2,048 time steps.
    \item Evaluation Episodes: To benchmark performance consistently, evaluation episodes were fixed to a duration of 2,000 seconds (200 control steps). 
\end{itemize}

The state space $S_t$ was designed to provide the agents with sufficient information to infer the dynamic flow state of the wind farm, including wake propagation delays. For each turbine $i$, the observation vector includes four primary signals: wind speed ($u_i$), wind direction ($\theta_i$), yaw misalignment angle ($\gamma_i$), and generated power ($P_i$).

The state vector was augmented with a history buffer. At each decision step, the agent observes a stack of the current measurement and the 10 preceding measurements. With a control update period of 10 seconds, this creates an effective observation window of 100 seconds. This allows the agent to better understand the time delay between an action imposed on an upstream turbine and the effect of the resulting wake advecting to or away from downstream turbines.



\section{Results}

This section presents the analysis results. In Section \ref{sec:training-method-results}, the agents identified within each training methodology are compared to one another to assess training progress. In Section \ref{sec:agent-behavior-results}, the behavior of the different agents is characterized. Time-series plots are presented for different protagonist-adversary pairs, and a cross-agent analysis is conducted using the best-performing agents from each method.

\subsection{Training Method Comparison}\label{sec:training-method-results}

This section compares the results of training the Arms Race, SSP, and Self-Play models. Each approach is compared by running each combination of protagonists, including the PyWake agent, and adversaries, including the clean and procedural noise cases, yielded within their respective training processes. \jq{Each protagonist-adversary pairing is evaluated using five standardized inflow episodes.} 
After these approaches individual results are presented, they are compared in line plots using different summaries of cross-agent analysis within each method\jq{, using the same five standardized inflow conditions}. 



\begin{figure}[thb!]
    \centering
    \includegraphics[width=0.85\linewidth]{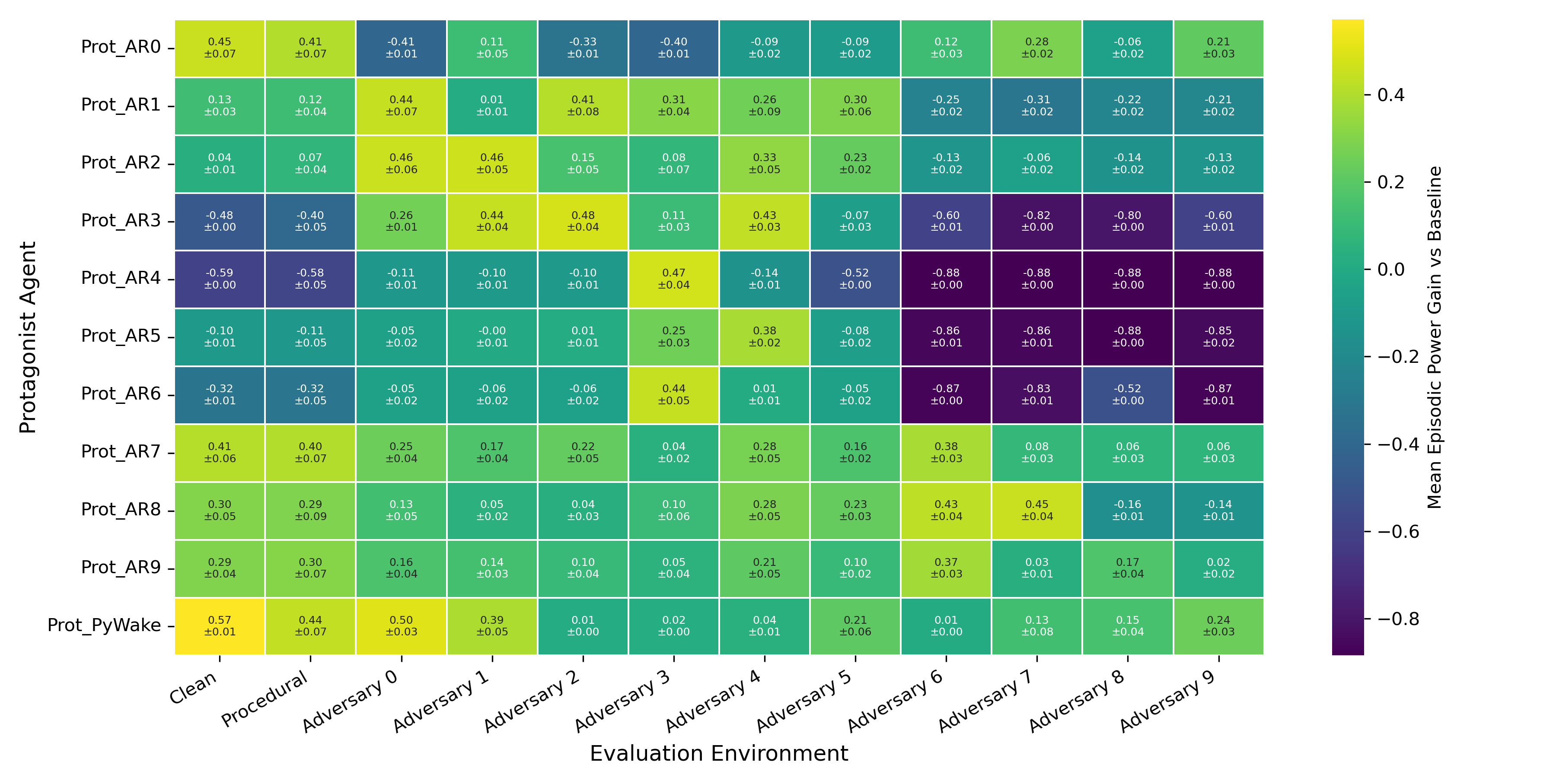}
    \caption{Arms Race training progress evaluation matrix. For each combination of considered protagonist and adversary, a standard set of trials is used to measure the performance of the wind farm system relative to the baseline no-steering/greedy operational case. The average increase over baseline operation, and the associated standard error, are reported.}
    \label{fig:arms_race_power_gain_heatmap}
\end{figure}

The Arms Race training results are summarized in Figure \ref{fig:arms_race_power_gain_heatmap}, which shows the reward of each trained protagonist operating within environments defined by clean, procedural, and adversarial conditions. These "gauntlet" evaluation results are averaged over five standardized inflow conditions in each case. 
As we increase the iteration number, the protagonist generally performs worse in the clean environment. This "catastrophic forgetting" can be observed in the protagonists trained in iterations 6 and 9, which effectively forgot how to operate in the procedural and clean environments, respectively. 
While the PyWake agent excels in the clean environment, as well as against adversary 0, it is otherwise vulnerable to being confounded by the adversarial agents, resulting in power very close to the baseline no-steering approach in four of the adversarial cases. 
Overall, the matrix reveals that the Arms Race adversaries consistently dominate their corresponding protagonists, with few protagonist iterations achieving positive rewards across a broad range of adversaries.

\begin{figure}[thb!]
    \centering
    \includegraphics[width=0.9\linewidth]{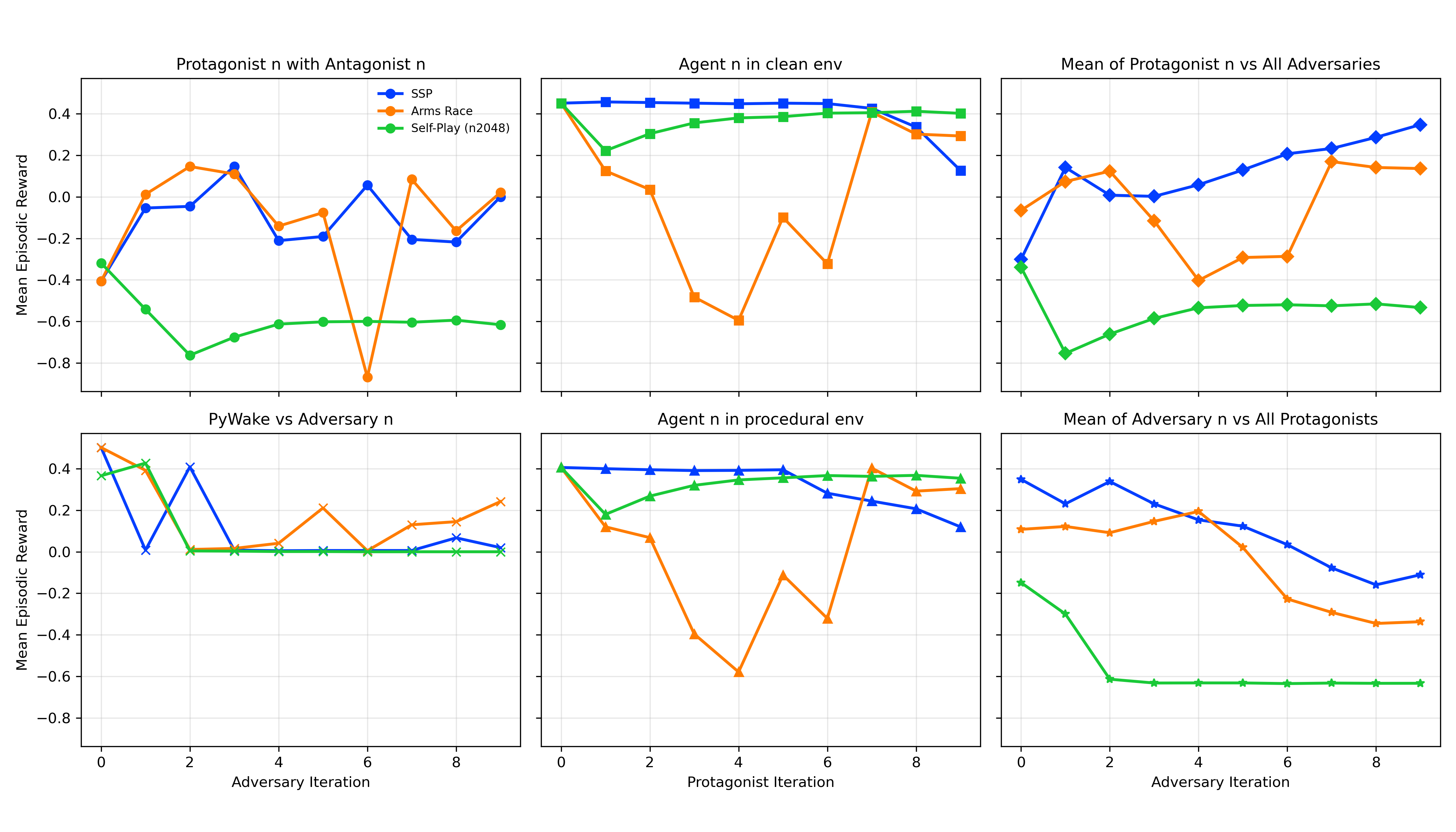}
    \caption{Summary plots comparing the different training approaches.}
    \label{fig:methods}
\end{figure}

We made similar heat maps to show the performance of the SSP, Arms Race, and Self-Play training strategies. Instead of directly comparing all three heat maps, it is more streamlined to compare different one-dimensional representations of them. This is done in Figure \ref{fig:methods}. 

The upper left plot in Figure \ref{fig:methods} compares the performance of protagonist $n$ versus adversary $n$. This corresponds to the values in the diagonal of the previous matrix. A plateau should indicate convergence. Based on these results, it appears that Self-Play has converged to a state where the adversary is dominating the protagonist. 
The Arms Race and SSP approaches do not seem converged and, in fact, it is likely that all examined training formulations will never converge. 

The lower left plot in Figure \ref{fig:methods} shows the performance of the expert PyWake agent operating alongside each trained adversary yielded by each method. Lower values indicate an adversary that is more formidable against this expert agent. 
The Self-Play approach yielded the strongest adversary.
The Arms Race and SSP adversaries suffer from catastrophic forgetting, drifting to states that do not confound the PyWake expert agent protagonist as severely as in the first iteration. 


The upper center plot in Figure \ref{fig:methods} shows the performance of protagonist $n$ when operating in a clean, adversary-free environment. Similarly, the lower center plot shows the performance of each protagonist while operating in an environment with procedural noise. Interestingly, the SSP approach yields the highest performance in nearly all iterations. After iteration 8, SSP forgets how to operate in a clean environment while Self-Play maintains its performance. The Arms Race approach shows a big dip in performance, reaching the lowest performance at iteration 4, corresponding to the system ``forgetting'' about the clean environment. The Arms Race controller then re-learns these important behaviors in subsequent iterations, as indicated by the rising performance after iteration 6. The Self-Play approach dips in performance after the first iteration and slowly recovers. 

The upper right plot in Figure \ref{fig:methods} shows the average performance of protagonist $n$ evaluated against all adversaries yielded by each considered training method. Similarly, the lower right plot shows the average performance of all protagonists yielded by the training method versus adversary $n$. Both of these metrics may be thought of as measures of convergence. The former is a measure of protagonist robustness, and the latter quantifies the robustness of adversaries. The Self-Play adversaries are by far the most potent using this measure.

\subsection{Agent Behavior Analysis}\label{sec:agent-behavior-results}

\begin{figure}[htb!]
    \centering
    \includegraphics[width=\linewidth]{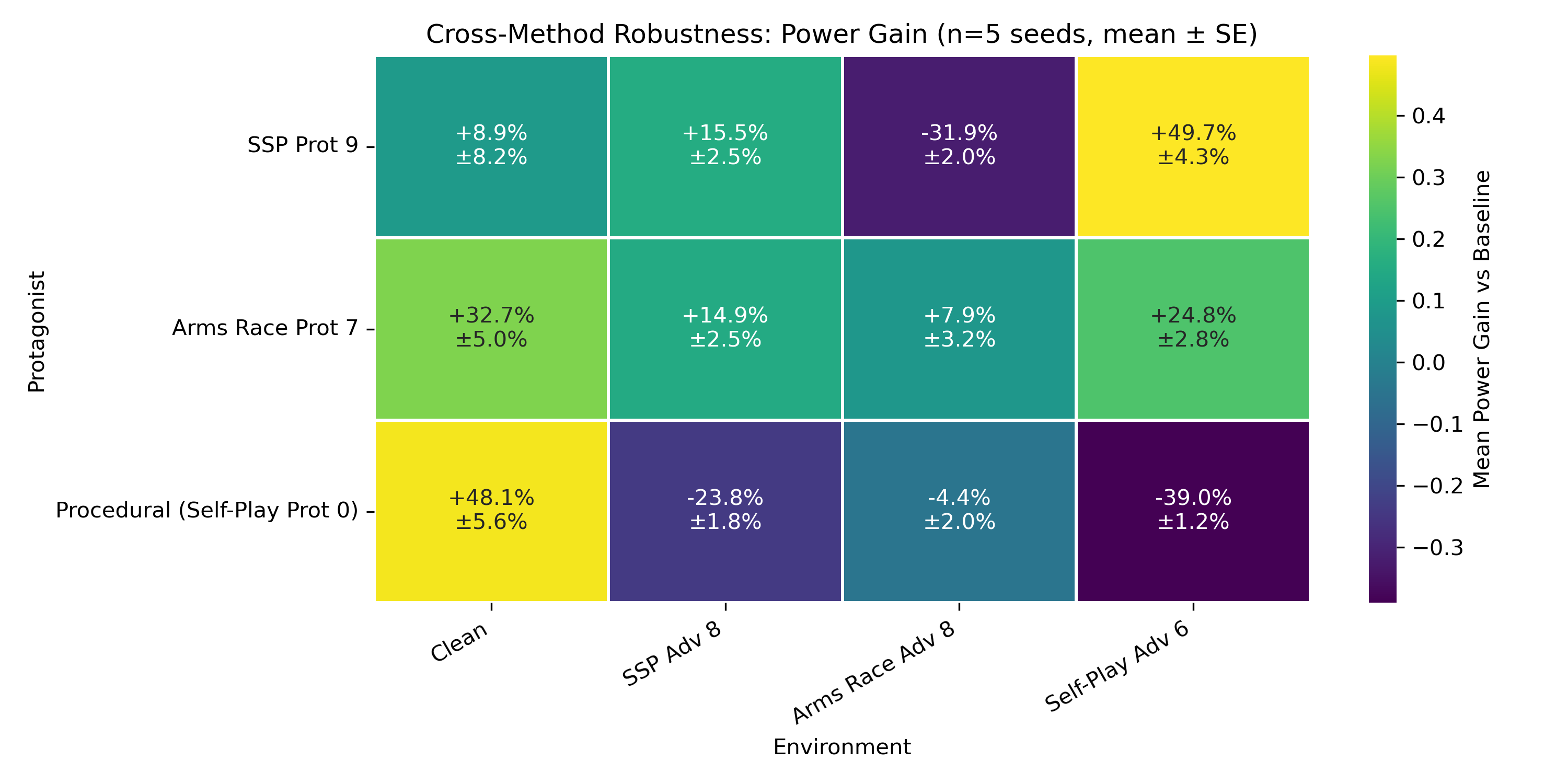}
    \caption{\label{fig:cross-compare}Cross-comparison results. The most formidable protagonists and adversaries identified in each training method, as well as a procedurally trained controller, are pitted against each other in five standard flow cases. Brighter colors show larger mean reward values.}
    
\end{figure}

Figure \ref{fig:cross-compare} shows a cross-comparison between the most formidable protagonists and adversaries, as well as a procedurally-trained protagonist, considering a standard set of five flow cases. The most powerful protagonist is identified as the one with maximum performance when averaged across all adversaries developed by the considered training method, corresponding to the maximum point of the three lines in Figure \ref{fig:methods}. We find that the Arms Race approach wins along both axis.  It produces by far the most robust protagonist and the most formidable adversary in cross-method comparison. It has a worst-case performance of a 8\% increase over baseline, compared to the other controller's worst-case performances of -32\% and -39\% power loss. 

\begin{figure}[htb!]
    \centering
    \includegraphics[width=0.45\linewidth]{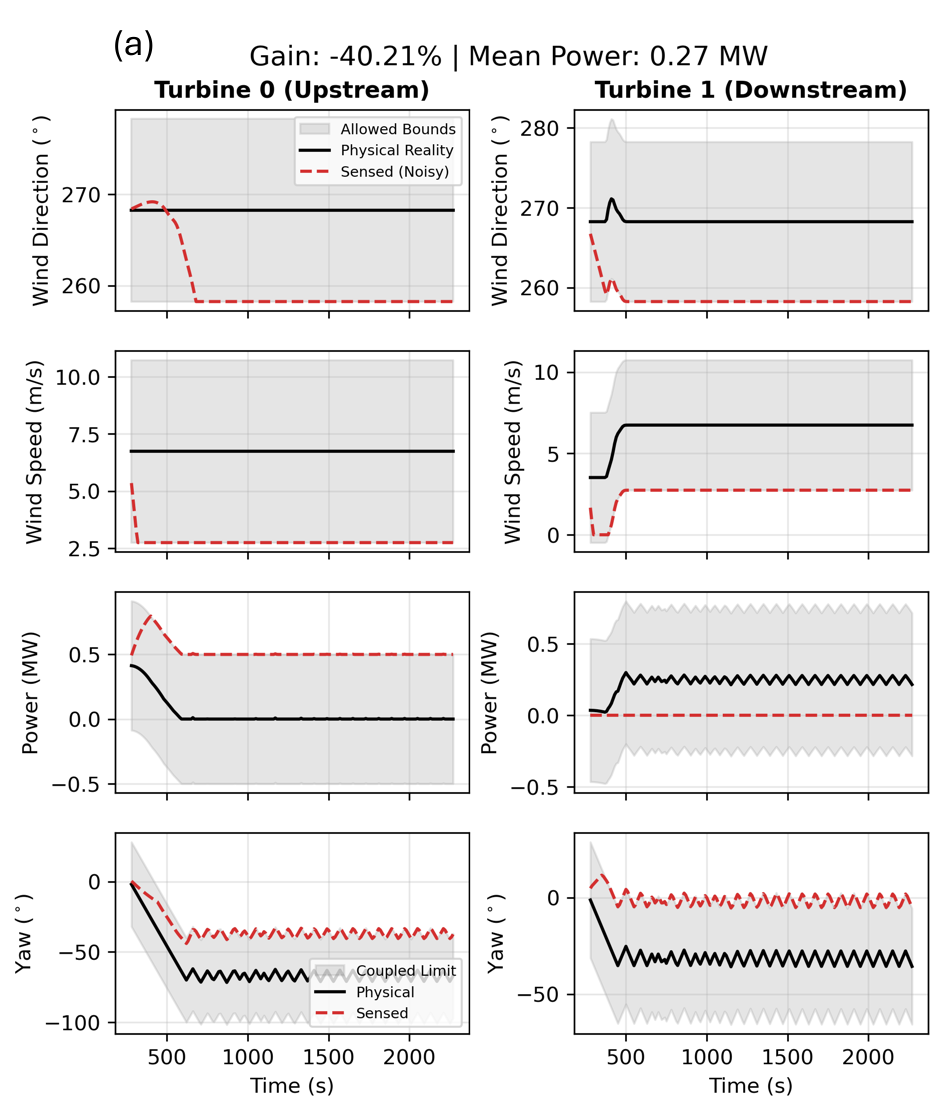}
    \includegraphics[width=0.45\linewidth]{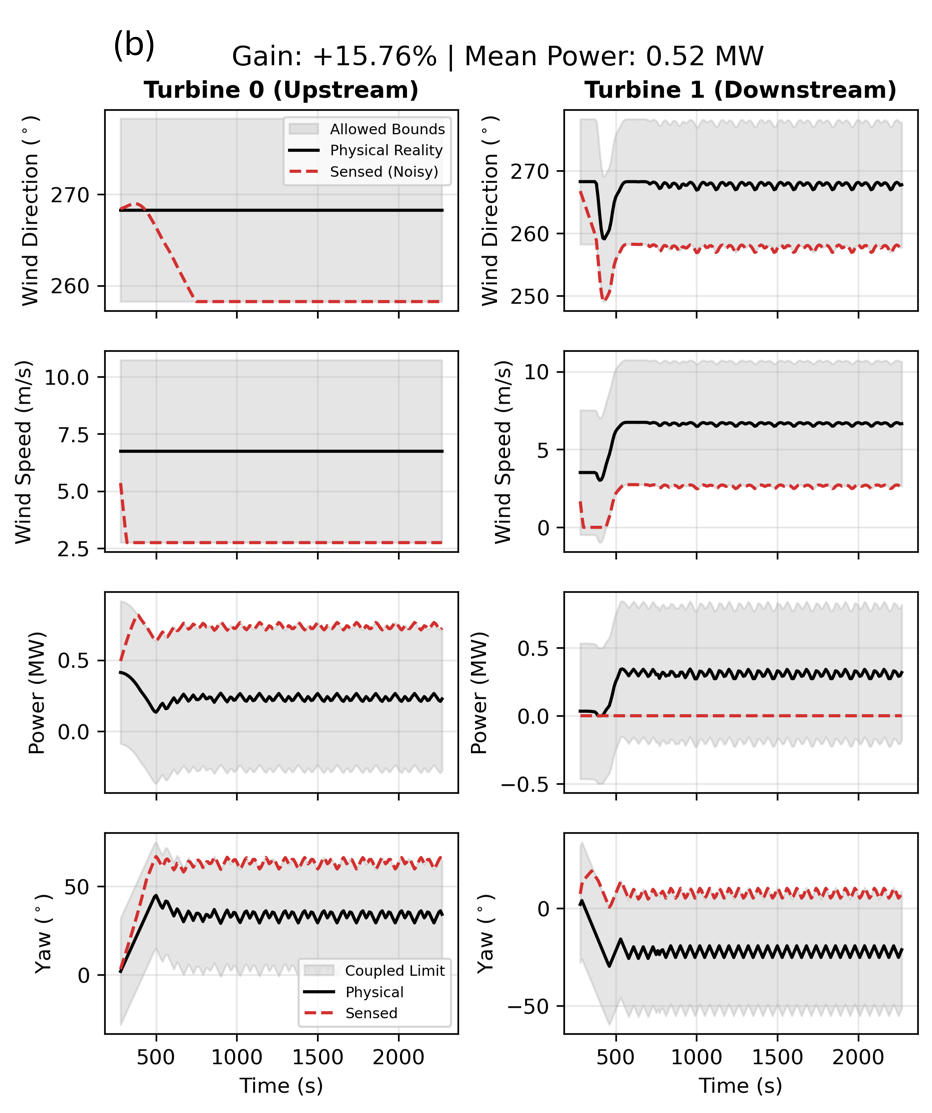}
    \caption{(a) Procedurally-trained protagonist agent versus Self-Play adversarial noise agent \#6 (b) Arms Race protagonist agent \#7 versus Self-Play adversarial agent \#6. The left columns show the upstream turbine and the right columns show the downstream turbine. The true and sensed wind direction, wind speed, power, and yaw are shown in dashed red and solid black lines, respectively. The grey clouds show the maximum error allowed within the adversarial environment and the maximum bias allowed in the procedural environment. }
    \label{fig:ppo-time-series}
\end{figure}

To understand how the Arms Race protagonist achieves this robustness, we compare the temporal behavior of agents under attack. 
In Figure \ref{fig:ppo-time-series}, the procedurally-trained protagonist is compared to Arms Race protagonist \#7 under adversarial attacks by Self-Play adversary \#6. When attacking the procedurally-trained agent, the adversary spoofs the back turbine power rating to read zero. This, alongside the other spoofed signals, completely confounds the procedurally-trained protagonist into calling for yaw offsets so extreme they would likely break the real machine. On the other hand, Arms Race protagonist agent \#7 is quite robust to the adversarial attacks, yielding 15.76\% more power than the no-steering baseline case. 


\section{Conclusions}

This study demonstrates that Arms Race adversarial training can produce wind farm controllers that are more robust than those trained with procedural noise. 
Under worst-case adversarial attack, the selected Arms Race controller maintained a +7.9\% power gain over baseline, while the procedurally-trained controller suffered a -39\% loss, performing worse than no steering at all.

It must be stressed that, while these results are averaged across different evaluation conditions, there is a great deal of seed-to-seed variability in the training conditions that will almost certainly yield different results on different training seeds. Further investigation is needed to find the best training architecture for this problem, particularly at scale. 

This study presents a framework for designing robust data-driven control strategies for wind power plants. As wind plant control pivots toward a central controller, it is increasingly likely that a malicious actor may hack into the control system, and a wind farm shutdown may be the intended effect. Furthermore, simulation training naturally assumes zero measurement errors, and this framework yields systems that are highly robust to potential sensor uncertainties. By designing wind farm control strategies that are robust to adversarial sensor biases, the security and real-world performance of wind power plants undergoing plant-level control may be improved.

This paper demonstrated that training a wind farm controller alongside adversarial agents that inject measurement errors, including temporally correlated noise, coordinated multi-sensor attacks, and strategically timed perturbations, produces more robust results than when training a controller alongside procedural noise.  

Establishing which adversarial training paradigm yields superior robustness in this canonical setting is a necessary prerequisite before scaling to larger farms, where computational costs would prohibit the extensive comparative analysis presented here. Future work will examine scalability to larger farm configurations \jq{and sim-to-real transfer strategies for field deployment}.


\ack
This work has been supported by the SUDOCO project, which receives funding from the European Union’s Horizon Europe Programme under the grant agreement No. 101122256.

\section*{Data availability}
The training scripts and intermediate results used to conduct this analysis are available here: https://doi.org/10.5281/zenodo.19037741

\bibliographystyle{iopart-num}
\bibliography{Torque26_references}

\end{document}